\documentclass[journal]{IEEEtran}
\usepackage{amsmath}
\usepackage{cite}
\usepackage{amsmath,amssymb,amsfonts}
\usepackage{threeparttable}
\usepackage{algorithmic}
\usepackage{graphicx}
\usepackage{textcomp}
\usepackage{stfloats}
\usepackage{booktabs}
\usepackage{array}
\usepackage{mathtools}
\usepackage{bm} %字体加粗
\usepackage{amsmath} %公式居中
\usepackage{amssymb}
\usepackage{colortbl}
\usepackage{color} % 给字体加颜色
\usepackage{float}
\usepackage{amsmath} % 花体字母
\usepackage{mathrsfs} %花体字母
\usepackage{makecell} % 表格内容居中
\usepackage{siunitx}
\usepackage{amssymb}
\usepackage{fancyhdr}
\usepackage{lastpage}
\usepackage{multirow}
\usepackage{threeparttable}

\begin{document}

\title{PUFA-GAN: A Frequency-Aware Generative Adversarial Network for 3D Point Cloud Upsampling}

\author{Hao~Liu,\IEEEmembership{}
        Hui~Yuan,~\IEEEmembership{Senior Member,~IEEE,}
        Junhui~Hou,~\IEEEmembership{Senior Member,~IEEE,}
        Raouf~Hamzaoui,~\IEEEmembership{Senior Member,~IEEE,}
        Wei~Gao~\IEEEmembership{Member,~IEEE,}

\IEEEcompsocitemizethanks{
\IEEEcompsocthanksitem This work was supported in part by the National Natural Science Foundation of China under Grants 62172259 and 61871342, the Taishan Scholar Project of Shandong Province (tsqn202103001), the Central Guidance Fund for Local Science and Technology Development of Shandong Province, under Grant YDZX2021002, the Hong Kong RGC under Grants CityU11202320, and the OPPO Research Fund.
\IEEEcompsocthanksitem Hao Liu and Hui Yuan are with the School of Control Science and Engineering, Shandong University, Jinan, 250061, China. \protect\\
E-mail: liuhaoxb@gmail.com, huiyuan@sdu.edu.cn
\IEEEcompsocthanksitem Junhui Hou is with the Department of Computer Science, City University of Hong Kong, Kowloon, Hong Kong, China. \protect\\
E-mail: jh.hou@cityu.edu.hk
\IEEEcompsocthanksitem Raouf Hamzaoui is with the School of Engineering and Sustainable Development, De Montfort University, Leicester, UK. \protect\\
E-mail: rhamzaoui@dmu.ac.uk
\IEEEcompsocthanksitem Wei Gao is with the School of Electronic and Computer Engineering, Peking University, and also with Peng Cheng Laboratory, China. \protect\\
E-mail: gaowei262@pku.edu.cn
\IEEEcompsocthanksitem Hao Liu and Hui Yuan contributed equally to this work; Hui Yuan and Junhui Hou are corresponding authors.
}}

%\thanks{Manuscript received xx xx, 2019; revised xx xx, 2019.}

\markboth{}%
{Shell \MakeLowercase{\textit{et al.}}: PUFA-GAN: A Frequency-Aware Generative Adversarial Network for 3D Point Cloud Upsampling}

\maketitle

\begin{abstract}
We propose a generative adversarial network for point cloud upsampling, which can not only make the upsampled points evenly distributed on the underlying surface but also efficiently generate clean high frequency regions. The generator of our network includes a dynamic graph hierarchical residual aggregation unit and a hierarchical residual aggregation unit for point feature extraction and upsampling, respectively. The former extracts multiscale point-wise descriptive features, while the latter captures rich feature details with hierarchical residuals. To generate neat edges, our discriminator uses a graph filter to extract and retain high frequency points. The generated high resolution point cloud and corresponding high frequency points help the discriminator learn the global and high frequency properties of the point cloud. We also propose an identity distribution loss function to make sure that the upsampled points remain on the underlying surface of the input low resolution point cloud. To assess the regularity of the upsampled points in high frequency regions, we introduce two evaluation metrics. Objective and subjective results demonstrate that the visual quality of the upsampled point clouds generated by our method is better than that of the state-of-the-art methods.
\end{abstract}


\IEEEpeerreviewmaketitle
\section{Introduction}
\IEEEPARstart{W}{ith} the progress of three-dimensional (3D) capture technology, 3D point clouds \cite{Li,Lv,Hu} have become widely used in many fields, such as immersive communication \cite{b1}, autonomous navigation \cite{b2}, etc. Point clouds captured by existing devices (e.g., LiDAR or depth cameras) are usually non-uniform and sparse, which negatively affects their subsequent processing. To address these issues, we propose a deep learning-based point cloud super-resolution method. Our method aims at solving the following ill-posed problem: given a non-uniform, sparse, noisy, low-resolution (LR) point cloud, find a uniform, dense, noise-free high-resolution (HR) point cloud which is consistent with the geometry distribution of the corresponding LR point cloud.

The mainstream point cloud upsampling methods can be roughly divided into two categories: optimization-based and deep learning-based approaches. Early optimization-based point cloud upsampling methods \cite{b3,b4,b5,b6,b7,b8,b9,Borges} rely more on the prior geometry knowledge of the input point clouds. Therefore, they can efficiently upsample flat regions. But for sharp regions (corner and edge) with a complex geometry distribution, their performance is severely limited. The existing deep learning-based methods \cite{b10,b11,b12,b13,b14,b15,b16,{bliu}} exploit convolutional neural networks. They mainly transfer the idea of image super-resolution to point clouds. However, all those methods mainly focus on how to generate a uniform HR point cloud and ignore the fundamental problem that the HR point cloud contains some noisy points. This severely affects the reconstruction quality, especially at the corner of the object (Fig. 1).

In this paper, we present a novel generative adversarial network (GAN) \cite{b17} for point cloud upsampling, called PUFA-GAN. In contrast to existing methods, our network addresses the important problem that the upsampled point cloud usually contains apparent noise. This noise affects the regularity of the generated shape and damages the local details of the generated HR point cloud (Fig. 1). To this end, we explicitly extract high frequency (HF) points (i.e., points extracted by a high-pass graph filter [see Section 3.3.1], such as points on a sharp corner or edge and noisy points) of the upsampled point cloud and send them to the discriminator directly so that the network can learn to recognize the implicit HF features and output a uniform upsampled point cloud with clean HF regions (i.e., the regions where the HF points are located).

\begin{figure*}[!htp]
\centering
\includegraphics[width=16cm,height=6.7cm]{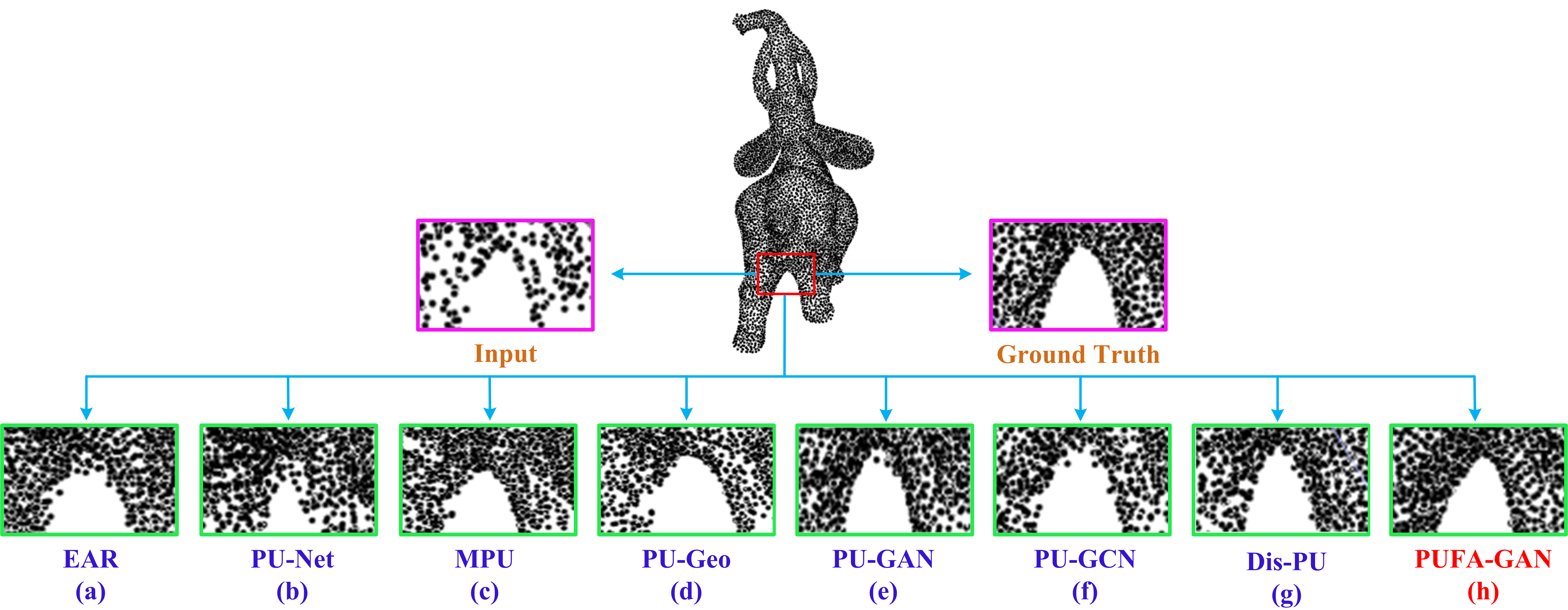}
\caption{Upsampling ($\bm{\times}$4) the non-uniform and sparse point cloud \textit{Elephant} with (a) EAR \cite{b6} (b) PU-Net \cite{b10}, (c) MPU \cite{b11}, (d) PU-Geo \cite{b13}, (e) PU-GAN \cite{b12}, (f) PU-GCN \cite{b15}, (g) Dis-PU \cite{b16} and (h) PUFA-GAN.}
\vspace{-1.5em}
\end{figure*}

GAN is hard to train and reach the Nash equilibrium with the generative learning. To balance the generation ability of the generator and the judgment accuracy of the discriminator, we design an efficient GAN to fully explore the global and local potential geometry characteristics of the point cloud. Specifically, we propose a dynamic graph hierarchical residual aggregation (DGHRA) unit and a hierarchical residual aggregation (HRA) unit for feature extraction and upsampling in the generator. DGHRA captures multiscale point-wise features by dynamically constructing a neighborhood graph, while HRA uses hierarchical residuals to capture the ample detail representation of the upsampled features, which generates natural geometric textures. Exploiting graph signal processing \cite{b18}, we leverage a graph filter (GF) \cite{b19} to extract HF geometry information (i.e., HF points) accurately. Next, we feed the HF points and the whole upsampled point cloud into the proposed frequency-aware discriminator (i.e., a global discriminator head and an HF discriminator head), respectively. In this way, the discriminator can learn both the global and HF features implicitly to better approximate the real target distribution. Finally, we devise an efficient identity distribution loss function to regularize the statistical distribution of the upsampled HR point cloud to be consistent with that of the input LR point cloud.

In summary, the key contributions of this paper are as follows.
\begin{itemize}
 \item We propose DGHRA and HRA units for point cloud feature extraction and expansion in the generator, respectively, to extract rich details by capturing hierarchical residual features.

 \item We extract HF points of the upsampled point cloud with a GF and propose a frequency-aware discriminator to better distinguish real HF points from fake ones produced by the generator. To the best of our knowledge, we are the first to apply a GF to a point cloud upsampling task.

 \item We introduce a simple and effective identity distribution loss to constrain the geometry distribution of the upsampled HR point cloud to be equal to that of the input LR point cloud.

 \item We propose two evaluation metrics (HF\_CD and HF\_HD) to assess the regularity of the upsampled point clouds in HF regions.
\end{itemize}

The remainder of this paper is organized as follows. Section II briefly reviews related work. Section III describes the proposed method in detail. Experimental results and conclusions are given in Section IV and V, respectively.

\section{RELATED WORK}

\textbf{Optimization-based upsampling methods}. In their pioneering work, Alexa \textit{et al}. \cite{b3} generated a Voronoi diagram on the moving least squares surface and interpolated points at the vertices of the corresponding diagram. A novel locally optimal projection (LOP) operator was introduced by Lipman \textit{et al}. \cite{b4} to robustly resample points and reconstruct surfaces. Preiner \textit{et al}. \cite{b5} proposed a continuous LOP for fast surface reconstruction. Huang \textit{et al}. \cite{b6} presented a progressive edge-aware (EAR) point cloud upsampling method, which first collects the points far away from the edge and then gradually upsamples the points to approach the edge singularity. By jointly optimizing the surface and inner points, Wu \textit{et al}. \cite{b7} proposed a deep points consolidation method to fill local holes caused by missing data. Recently, Dinesh \textit{et al}. \cite{b8} proposed a model-based point could super-resolution method via graph total variation on surface normal. In \cite{b9}, the same authors also proposed a fast graph total variation method for color point cloud super-resolution. Borges \textit{et al}. \cite{Borges} proposed an efficient point cloud super-resolution via self-similarity of the point cloud.

\begin{figure*}[!htp]
\centering
\includegraphics[width=16.8cm,height=11cm]{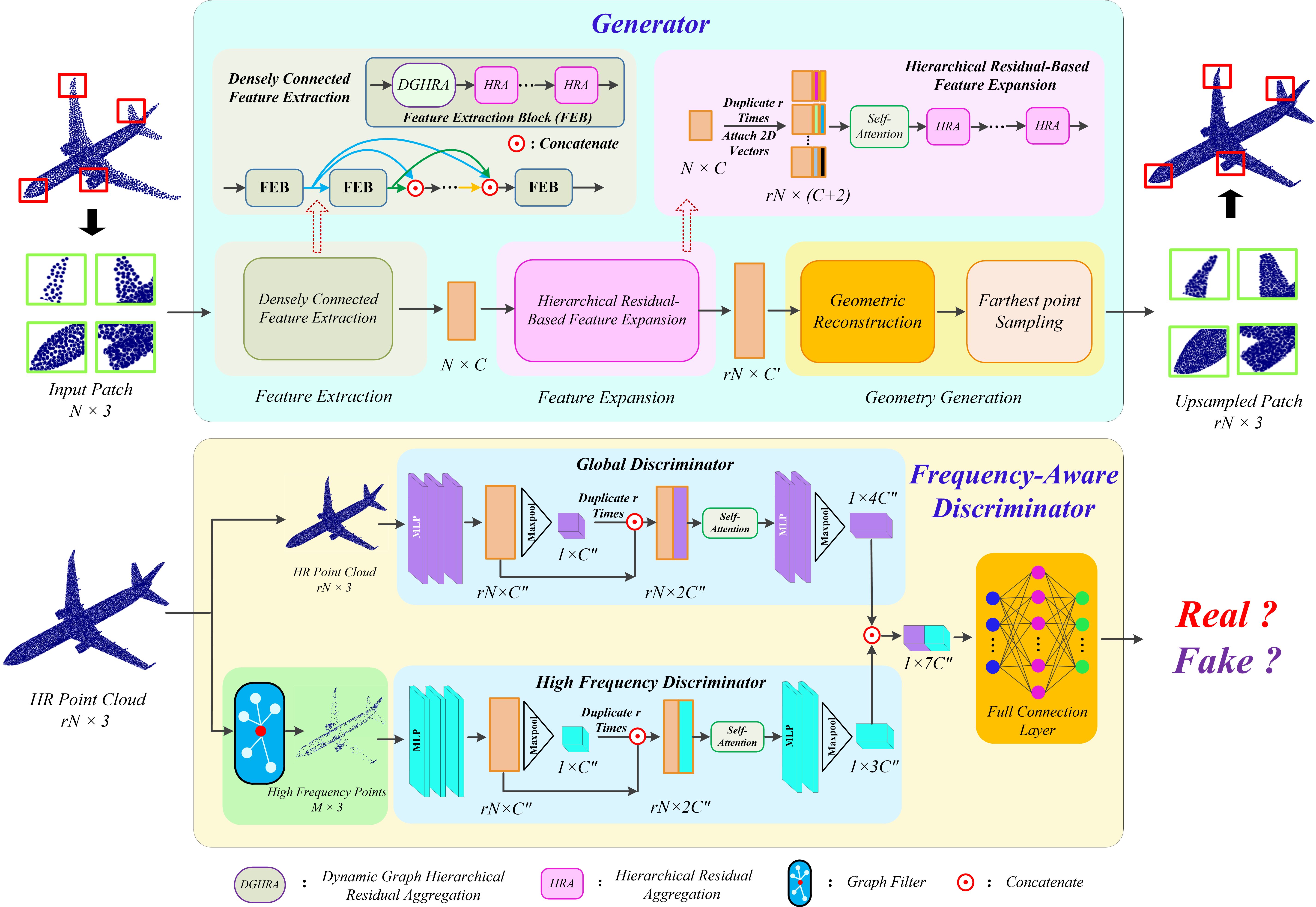}
\caption{Architecture of PUFA-GAN. The proposed network consists of two parts: a generator and a discriminator. For an arbitrary LR input point cloud patch with $N$ points, the generator aims to output the corresponding HR point cloud patches with $rN$ points ($r$ is the upsampling ratio), and the discriminator determines where the HR point cloud patches come from (i.e., the output of the generator or the ground truth). Specifically, the feature extraction module first learns point-wise descriptive features. Then the upsampled features with rich details can be obtained by the feature expansion module. Finally, the geometry generation module maps the upsampled features back to the geometry domain and outputs HR point cloud patches. The proposed frequency-aware discriminator guides the generator to produce more realistic upsampling point clouds by global and HF contexts.}
\vspace{-0.5em}
\end{figure*}

\textbf{Deep Learning-based upsampling methods}. PointNet \cite{b20}, a pioneering point-based deep learning model, uses the permutation invariance of a point cloud to extract global features by using multilayer perceptrons (MLPs). However, MLPs cannot capture local features well. PointNet++ \cite{b21} further uses a set abstraction layer to efficiently collect local presentation by the ball query method. Compared with optimization-based upsampling methods, data-driven methods have shown impressive performance. EC-Net \cite{b22} is a deep learning-based upsampling method that aims at generating points near the edges and on the underlying surface, while preserving the sharp features. Yu \textit{et al}. \cite{b10} introduced an upsampling network called PU-Net which extracts multiscale features by PointNet++ and expands them with shared MLPs. Later, Wang \textit{et al}. \cite{b11} proposed a multiscale upsampling method (MPU), which progressively achieves efficient upsampling operations by controlling the upsampling factor ($\times $2) for hierarchical detail reconstruction. But the progressive feature increased the complexity of the network and had potential error accumulation. Qian \textit{et al}. \cite{b13} used a local surface parameterization to represent a 3D point cloud in a 2D parametric domain and then used the normal vectors to generate the upsampled 3D point cloud adaptively from the 2D parametric surface. However, the method requires the normal vectors which are usually not available from the raw point clouds directly. In \cite{b14}, the same authors proposed an adaptive point cloud interpolation method based on linear estimation theory. Qian \textit{et al}. \cite{b15} proposed a graph convolutional network (GCN)-based point cloud upsampling method (PU-GCN) which includes an inception dense GCN feature extraction and a node shuffle-based feature expansion module. Li \textit{et al}. \cite{b16} proposed two cascaded sub-networks (Dis-PU) for point cloud upsampling. In this method, the dense generator first produces a coarse upsampled point cloud. Then, the spatial refiner restores a fine upsampled point cloud from the coarse one by predicting the point-wise spatial offsets. Liu \textit{et al}. \cite{bliu} proposed a progressive geometry upsampling method with adversarial learning. Metzer \textit{et al}. \cite{b23} proposed a point cloud consolidation method (Self-Net) with self-supervision learning\cite{Doersch,Gidaris}, which uses the curvature and density of the point cloud to generate sharp edge points or points in sparse regions. Although both our method and Self-Net use HF points, the optimization objectives are different. Self-Net pays more attention to geometry sharpness by forcing more points to be close to the HF regions. This may affect the uniformity of the upsampled point cloud, especially for a sparse point cloud. In contrast, PUFA-GAN only limits the noise in HF regions by distinguishing HF points, and does not push points to the HF regions. Hence, PUFA-GAN not only maintains the global uniformity, but also generates clean edges in HF regions.

Li \textit{et al}. \cite{b12} proposed an adversarial learning network (PU-GAN) for point cloud upsampling. They introduced an efficient global discriminator to accurately distinguish the upsampled point clouds of the generator from the ground truth. They also promoted the generator to produce upsampled point clouds that are consistent with the distribution of the ground truth through adversarial learning. However, their simple global discriminator tends to ignore the subtle description features of the upsampled point cloud, leading to local noise and detail distortions.
Compared to PU-GAN, i) we propose a novel DGHRA-based densely connected feature extraction and a cascaded HRA-based feature expansion module, which makes full use of multi-scale residual information to obtain more expressive features; ii) we propose an efficient frequency-aware discriminator to suppress noise in HF regions; iii) in addition to using the general loss functions in PU-GAN, we propose an identity distribution loss function which reduces the solution space of the network and accelerates its convergence.

\begin{figure}[h]
\centering
\includegraphics[width=8.9cm,height=4.2cm]{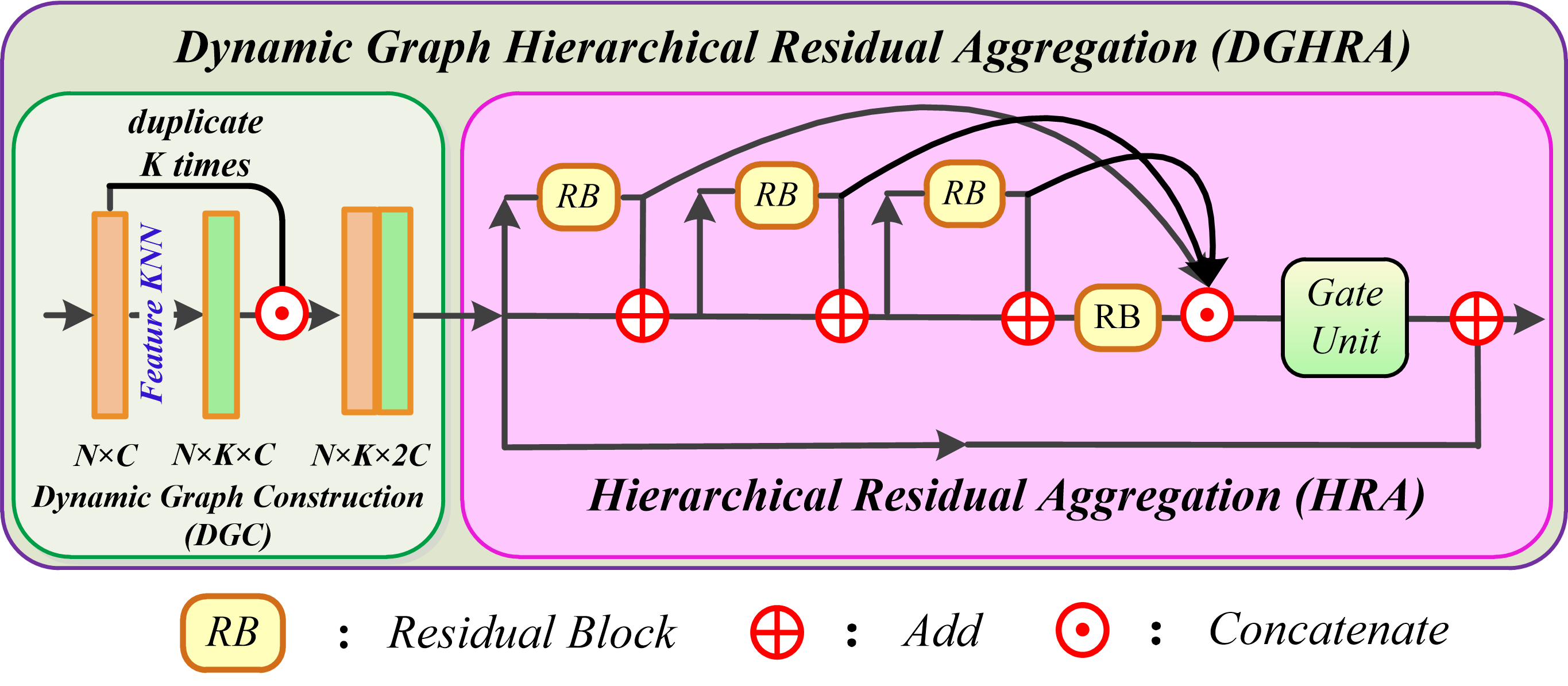}
\caption{Proposed DGHRA and HRA units.}
\end{figure}

\vspace{1.5em}
\section{METHOD}
\subsection{Overview}

In general, a point cloud is a set of points given by their 3D Cartesian coordinates (geometry information), together with a number of attributes such as color and reflectance. In this paper, we only consider the geometry information. Thus, a point cloud consisting of $N$ points can be represented by a matrix ${P} \in$ ${\mathbb{R}^{{N} \times 3}}$. Given a non-uniform, sparse point cloud ${P_{ori}} \in$ ${\mathbb{R}^{{N} \times 3}}$, the aim of the proposed method is to output a uniform, dense upsampled point cloud ${P_{up}} \in$ ${\mathbb{R}^{{rN} \times 3}}$, where $r$ denotes the upsampling ratio. Simultaneously, ${P_{up}}$ should satisfy the following conditions: 1) ${P_{up}}$ can generate the same underlying surface as ${P_{ori}}$. 2) the points in ${P_{up}}$ are evenly distributed on the underlying surface. 3) ${P_{up}}$ should contain rich geometry details, especially in the HF regions.

Recently, GANs have been very successful in 2D image processing \cite{b24,b25} and 3D object generation \cite{b26,b27}. For upsampling, we propose PUFA-GAN as shown in Fig. 2, where the generator produces an upsampled point cloud ${P_{up}}$, and the frequency-aware discriminator is used to identify a fake (the output of the generator) ${P_{up}}$ efficiently. Note that PUFA-GAN takes patches instead of the whole point cloud as the processing unit. In the following, unless otherwise specified, the mentioned point clouds (e.g., ${P_{ori}}$, ${P_{up}}$) denote patches.

The remainder of this section is organized as follows. The detailed framework of the proposed generator is described in sub-section 3.2. The novel frequency-aware discriminator is proposed in sub-section 3.3. Finally, a joint loss function for end-to-end training is introduced in sub-section 3.4.

\subsection{Generator}
The proposed generator includes three modules: feature extraction, feature expansion, and geometry generation, as shown in Fig. 2 (top).

\subsubsection{Feature Extraction}
The feature extraction module extracts point-wise representative features $F \in$ ${\mathbb{R}^{{N} \times C}}$ ($C$ is the number of features) of the input point cloud ${P_{ori}} \in$ ${\mathbb{R}^{{N} \times 3}}$. DenseNet \cite{b28} has strong feature extraction capabilities due to its dense connections. For each network layer, the features of all preceding layers are used as inputs, and its own features are also used as inputs to all subsequent layers. Inspired by this idea, we propose a densely connected feature extraction module to connect all previous multiscale features (Fig. 2 top left). Specifically, each feature extraction block (FEB) consists of the concatenation of one DGHRA and several HRAs. The details of DGHRA and HRA are as follows.

\textbf{DGHRA and HRA Units}. How to extract expressive point-wise features is a critical task for the network. DGCNN \cite{b29} can effectively process point clouds through a dynamic neighborhood feature graph for ``local'' feature collection. However, the original DGCNN cannot discover fine differences between features. The residual features \cite{b30} have been shown to be very effective for super-resolution tasks \cite{b31,b32}. Classical stacked residual blocks can efficiently extract subtle features dissimilarity. But the residual features are fused with the identity features before propagation to the next residual block. As a result, the subsequent residual blocks cannot explore pure residual features individually, which deteriorates the network performance. RFANet \cite{b33} effectively alleviates the above problem by explicitly extracting pure residual features.

Inspired by DGCNN and RFANet, we present a DGHRA unit for feature extraction. The unit can be regarded as a combination of dynamic graph construction (DGC) and HRA. It not only constructs a multiscale feature graph based on neighborhood information dynamically but also captures hierarchical pure residuals from each residual block for fine grained representation.

As shown in Fig. 3, DGC first forms each point feature ${\hat f_{i}}$ from a local neighborhood that is computed dynamically via k-nearest neighbor (KNN) search based on feature similarity, i.e., \begin{equation} {\hat f_i^j} = [{f_i} \odot ({f_i} - {f_j})], \end{equation} where ${\hat f_i^j}$ is the feature gradient of the $i$-th current point and its $j$-th feature neighbor, ${f_i}$ is the $i$-th current point feature, ${f_j}$ is the $j$-th feature neighbor of ${f_i}$, and $\odot$ is the concatenation operation.

Then the proposed HRA module is applied to obtain more elaborate features for each point. Specifically, HRA includes four residual blocks (RBs) (Fig. 4 (a)) and one gate unit (Fig. 4 (b)). It independently extracts the pure residual features of each RB, which are concatenated together with the output of the last RB. Next, a gate unit is used to recalibrate the fused hierarchical residuals before element-wise addition with the identity feature. In this way, the role of the residuals is brought into full play, which allows the network to fully explore the nuances among features. Finally, a max pooling operation is applied to achieve point-wise feature sublimation by finding the maximum point-wise feature response of all nearest neighbors. It is worth noting that we also use cascaded HRA units in the feature expansion module of the generator.

\textbf{The RB} can help preserve more details and accelerate network convergence. As shown in Fig. 4 (a), it contains three linear layers with a rectified linear unit (ReLU). To build a lightweight network, we use bottleneck layers to reduce the number of trainable convolution kernels.

\textbf{The gate unit} recalibrates the fused features. The proposed gate unit is composed of SE-Net \cite{b34} and a linear layer. Because there are many redundancies among fused hierarchical residuals, we use the efficient channel attention mechanism (SE-Net, see Fig. 4 (b)) to learn global information by global average pooling to selectively stress significant features and suppress invalid ones. Then, we use a linear layer to refine and align the channels of fused residuals for subsequent element-wise addition.

\begin{figure*}[!htp]
\centering
\includegraphics[width=13.7cm,height=4.1cm]{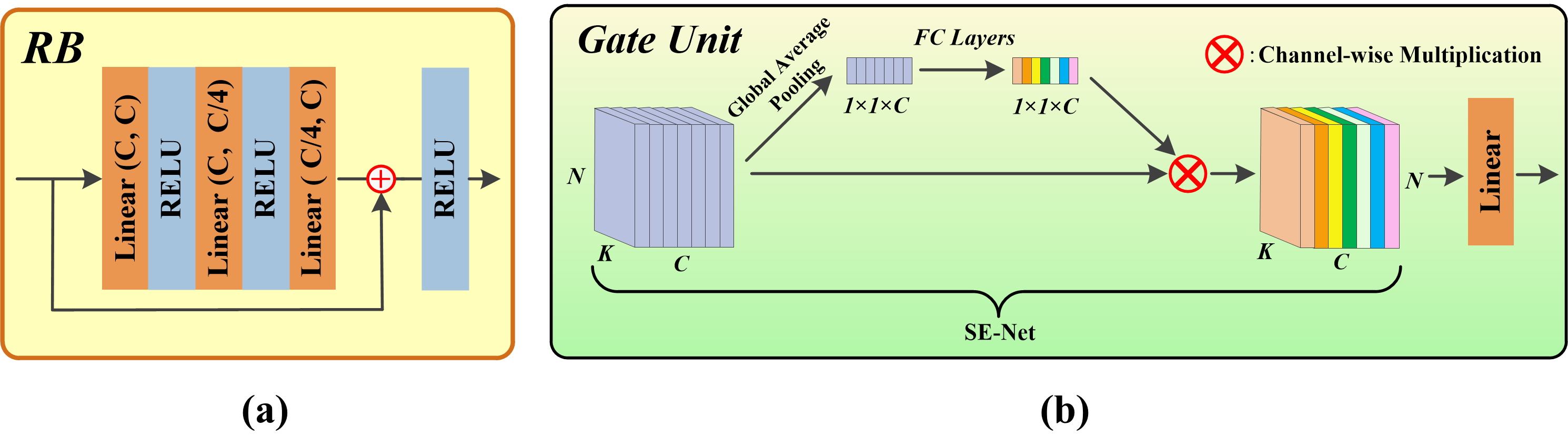}
\caption{Overview of the proposed residual block (RB) (a) and gate unit (b), where Linear (C1, C2) indicates a linear layer, and C1 and C2 are the dimension of the input and output feature vector.}
\end{figure*}

\subsubsection{Feature Expansion}
Feature expansion aims to expand the feature vector $F \in$ ${\mathbb{R}^{{N} \times C}}$ to the upsampled feature vector $\hat F \in$ ${\mathbb{R}^{{rN} \times C^{\prime}}}$,  where $C^{\prime}$ is the number of features in the upsampled points. It uses the local smoothing property of the point cloud to generate rich upsampled features. Specifically, we first duplicate feature vector $F \in$ ${\mathbb{R}^{{N} \times C}}$ $r$ times to get the upsampled feature vector $\tilde F \in$ ${\mathbb{R}^{{rN} \times C}}$. To increase the diversity of $\tilde F$, we use the 2D grid representation strategy proposed in FoldingNet \cite{b35}, i.e., we attach a 2D vector to $\tilde F$ to make each upsampled point of $\tilde F$ unique, as shown in Fig. 2 (top). Next, we send the upsampled feature vector with dimension $rN \times (C + 2)$ to the proposed cascaded HRA units (Fig. 2 (top)) to obtain the upsampled fine feature vector $\hat F \in$ ${\mathbb{R}^{{rN} \times C^{\prime}}}$.

\subsubsection{Geometry Generation}
The geometry reconstruction module uses several MLPs to map $\hat F$ from the implicit feature domain back to the explicit geometry domain (i.e., upsampled 3D geometry coordinates ${P_{up}} \in$ ${\mathbb{R}^{{rN} \times 3}}$ ). To ensure the uniformity of the upsampled point cloud, we follow the strategy in \cite{b12}. First, the feature expansion module generates $(r+2)N \times 3$ point features. Then, farthest point sampling (FPS) is used to obtain the final upsampled point cloud with size $rN \times 3$.

\subsection{Frequency-Aware Discriminator}
The role of the discriminator is to identify where ${P_{up}}$ comes from, i.e., the output of the generator (Fake) or the ground truth (True)? The discriminator can force the generator to produce ${P_{up}}$ which approximates the latent distribution of the target point cloud. To identify ${P_{up}}$ efficiently, the work in \cite{b12} adopts a basic encoder network \cite{b36} as the backbone of the discriminator. However, when the local density of the input point cloud is large (i.e., the local feature similarity is very high), a single global discriminator cannot recognize the local details well, especially for the HF regions which usually contain noise. In other words, it allows point clouds with local noise to easily fool the discriminator and then distracts the generator when generating the HF points. Therefore, the generator will tend to produce irregular HF regions with noise, which severely affects the upsampling performance (Fig. 1(e)).

\begin{figure}[h]
\centering
\includegraphics[width=8.3cm,height=5.3cm]{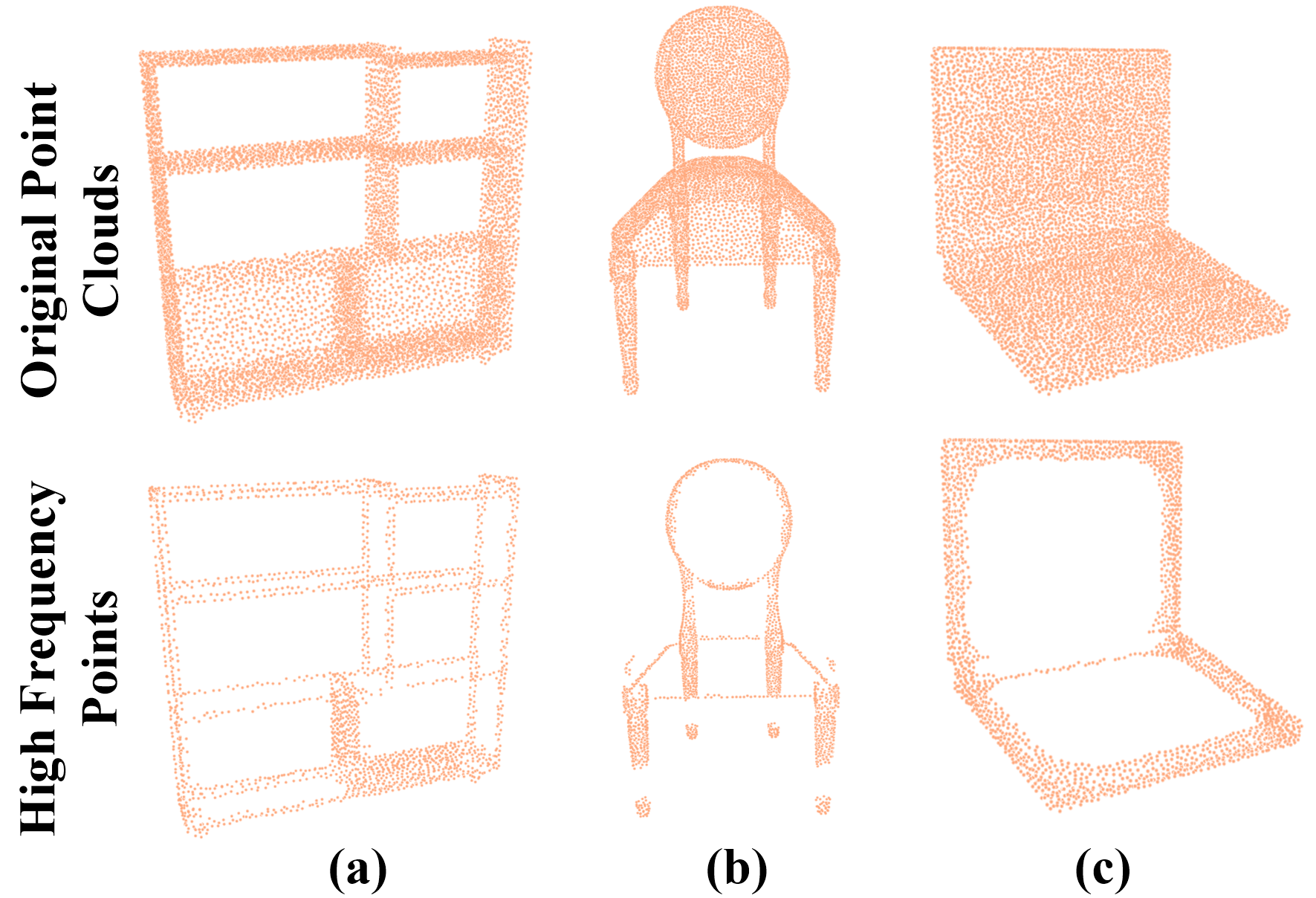}
\caption{High frequency points extraction by graph filter. The top row corresponds to original point clouds, and the bottom row shows the high frequency points after graph filtering.}
\end{figure}

To obtain a uniform point cloud with faithful HF regions, we propose a novel frequency-aware discriminator, as shown in Fig. 2 (bottom). Specifically, for each input ${P_{up}} \in$ ${\mathbb{R}^{{rN} \times 3}}$, a graph filter is first used to extract the HF points ${P_{b}} \in$ ${\mathbb{R}^{{M} \times 3}}$ (see Fig. 5) explicitly, where $M (M < rN)$ denotes the number of HF points. Then, ${P_{up}}$ and ${P_{b}}$ are fed into the proposed global and HF discriminator, respectively, for more accurate discriminative results. The former considers the global context for better skeleton recognition, while the latter suppresses noise in HF regions by mining the differences of local HF features.

\subsubsection{HF Points Extraction Using Graph Filter}
\textbf{Graph construction}. For an arbitrary point cloud $\bm{P} \in$ ${\mathbb{R}^{{K} \times 3}}$  (with $K$ points), a graph $\bm{G}$ can be constructed as follows \begin{equation} \bm{G} = \{ \bm{X} = \{ {x_1},{x_2},...,{x_K}\} ,\bm{A}\}, \end{equation} where $\bm{X}$ is the set of nodes, node ${x_k}$ ($k = 1,...,K$) refers to the $k$-th point in the point cloud, and $\bm{A} \in$ ${\mathbb{R}^{{K} \times K}}$ is the adjacency matrix indicating the edge dependencies between the points. We define $\bm{A}$ as follows:  \begin{equation}  {A_{i,j}} = \left\{ {\begin{array}{*{20}{c}}
{f({{\left\| {{x_i} - {x_j}} \right\|}_2}), {\rm{  }} if  {{\left\| {{x_i} - {x_j}} \right\|}_2} < \varepsilon }\\
{\qquad 0,  \qquad \qquad otherwise ,}
\end{array}} \right.
\end{equation} where $f( \cdot )$ is a Gaussian function, and $\varepsilon$ is a threshold. Note that $\bm{A}$ is a symmetric matrix.

\textbf{Graph filter theory}. Let the adjacency matrix $\bm{A}$ be a graph shift operator, i.e., it converts a signal $\bm{s} \in$ ${\mathbb{R}^{{K} \times 1}}$ (an attribute or feature of point clouds) into a shifted signal $\bm{y}=\bm{A}\bm{s}$. The linear, shift-invariant all-pass graph filter $\bm{H(A)}$ is defined as a polynomial of the graph shift operator \cite{b19}, i.e., \begin{equation} \bm{H(A)} = \sum\nolimits_{l=0}^{L - 1} {{h_l}{\bm{A}^l} = {h_0}I + {h_1}\bm{A} +  \cdot  \cdot  \cdot  + {h_{L - 1}}{\bm{A}^{L - 1}}}, \end{equation} where $L$ is the number of filter taps, and ${h_l}$ is the $l$-th filter coefficient. Hence, the output of the all-pass graph filter can be further written as $\bm{\tilde y}=\bm{H(A)}\bm{s}$. To explore the frequency representation of graph filters, the following eigen decomposition \cite{b18} is considered: \begin{equation} \bm{H(A)}= \bm{V} \bm{H(\Lambda)} \bm{V^{-1}}, \end{equation} where $\bm{V} \in$ ${\mathbb{R}^{{K} \times K}}$ is the eigenvector matrix of $\bm{H(A)}$, $\bm{H(\Lambda)} \in$ ${\mathbb{R}^{{K} \times K}}$ is the diagonal eigenvalue matrix with ordered eigenvalues ${\lambda _1} \ge {\lambda _2} \ldots  \ge {\lambda _K}$ (there is a one-to-one correspondence between the ordered eigenvalues and the graph frequencies, i.e., the larger the eigenvalue, the lower the frequency, and vice versa \cite{b18}), and $\bm{V^{-1}} \in$ ${\mathbb{R}^{{K} \times K}}$ is the corresponding graph Fourier transform (GFT) basis.
Therefore, the all-pass graph filter can be described as:
\begin{equation}
\begin{array}{l}
\bm{\tilde y} = \bm{H(A)s}\\
  \ \ \ = \bm{VH(\Lambda )}\underbrace {{\bm{V^{ - 1}}}\bm{s}}_{GFT}\\
  \ \ \ = \bm{V}\underbrace {diag[H({\lambda _1}),H({\lambda _2}),...,H({\lambda _K})]\bm{\tilde s}}_{Filtering \; in \; graph \; Fourier \; domain}\\
  \ \ \ = \underbrace {\bm{V}diag{{[H({\lambda _1}){{\tilde s}_1},H({\lambda _2}){{\tilde s}_2},...,H({\lambda _K}){{\tilde s}_K}]}^T}}_{Inverse \; GFT},
\end{array}
\end{equation}
i.e., the input signal $\bm{s}$ is first transformed into the graph Fourier domain signal $\bm{\tilde s}$ (i.e., $\bm{V^{-1}s}$) by GFT. Then, $\bm{\tilde s}$ is filtered by the frequency response ${diag[H({\lambda _1}),H({\lambda _2}),...,H({\lambda _K})]}$ of $\bm{H(A)}$, where $H({\lambda _k}) = \sum\nolimits_l^{L - 1} {{h_l}\lambda _k^l}$ \cite{b18} is the frequency response of ${\lambda_k}$. Finally, an inverse GFT transforms the filtered signal back into the graph node domain.

\textbf{Extracting HF points by graph filter}. Low frequency components of the point cloud usually correspond to smooth regions (flat areas), while high frequency components correspond to HF regions (edge, corner). In this paper, we aim to find a filter on the graph $\bm{{G}=\{X, {A} \}}$ to precisely extract HF points for an arbitrary point cloud. Therefore, we adopt a simple yet effective Haar-like high-pass graph filter \cite{b37}. The polynomial of the graph filter can be expressed as \begin{equation} \begin{array}{l}
\bm{H({A})} = \bm{I - {A}}\\
 \qquad \ \ = \bm{VH({\Lambda} ){V^{ - 1}}}\\
 \qquad \ \ = \bm{V}\left[ {\begin{array}{*{20}{c}}
{1 - {{ \lambda }_1}}&0& \cdots &0\\
0&{1 - {{\lambda }_2}}&{ \cdot  \cdot  \cdot }&0\\
 \vdots & \vdots & \ddots & \vdots \\
0&0&{ \cdot  \cdot  \cdot }&{1 - {{ \lambda }_K}}
\end{array}} \right]\bm{{V^{ - 1}}},
\end{array} \end{equation} where the frequency response of the graph filter is $H({\lambda _i}) = 1-{\lambda _i}$. Because the eigenvalues of $\bm{A}$ are arranged in descending order (i.e., ${\lambda _1} \ge {\lambda _2} \ldots  \ge {\lambda _K}$), $1-{\lambda _i} \le 1-{\lambda_{i+1}}$. Therefore, by setting the smaller eigenvalues of $\bm{H({A})}$ to zero, the low frequency parts are suppressed and high frequency parts are amplified after applying the high-pass graph filter.

Given $\bm{P} \in$ ${\mathbb{R}^{{K} \times 3}}$, we can apply $\bm{H({A})}$ to get a filtered point cloud $\bm{H({A})P}$. Specifically, because $\bm{H({A}) = I - {A}}$, the response of the $i$-th point $x_i$ of the point cloud $\bm{P}$ can be formulated as: \begin{equation} \bm{(H({ A})P)}_i = {x_i} - {\sum\nolimits_{j=1}^K {A}_{i,j}}{x_j}, \end{equation} In fact, $\bm{(H({A})P)}_i$ is equivalent to the feature discrepancy between the current point and the linear combination of its neighbors, which reflects the variation between the current point and its neighbors. For example, when the variation becomes large, the Euclidean distance between the current point and its neighbors increases, meaning that the current point has drastically mutated and can be regarded as an HF point.

Eventually, we can obtain the variation amplitude of each point $x_i$ ($i=1,...,K$) by calculating the $L2$-norm of $\bm{(H({ A})P)}_i$ in Eq.(8). A point with a larger $L2$-norm has more HF local variations and is an HF point candidate. By reordering the rows of the point cloud  $\bm{P} \in$ ${\mathbb{R}^{{K} \times 3}}$ in descending order according to the $L2$-norm from Eq.(8), we select the top $M$ (empirically set to 256) points as HF point cloud $\bm{P_{b}} \in$ ${\mathbb{R}^{{M} \times 3}}$.

\subsubsection{Discriminator structure}
As shown in Fig. 2 (bottom), the proposed frequency-aware discriminator includes two heads (a global and an HF head), where each head has the same backbone \cite{b12}. In particular, the input goes through two MLP layers (32 and 64 output feature dimensions) with max pooling to get the global (resp. HF) response vector, and the duplicated global (resp. HF) response vector is concatenated with the output of the previous MLPs for feature enhancement. Then, a self-attention module \cite{b38} is applied to strengthen the feature interaction among points. Again, two MLP layers (128 and 256 output feature dimensions) with max pooling are used to acquire the corresponding global (resp. HF) feature response vector. Finally, the feature response of each head is concatenated to form the final discriminant vector, and the confidence value of the discriminator is generated by three fully connected (FC) layers. Note that we assume the confidence value of the ground truth (True) is close to 1 and the confidence value of the output (Fake) of the generator is close to 0.

\subsection{Loss Functions}
We propose a joint loss to train PUFA-GAN in an end-to-end fashion.

\textbf{Adversarial loss}. We adopt the least-squared adversarial loss \cite{b39} to train the generator $G$ and discriminator $D$,
\begin{equation}
{L_{gen}}({P_{up}}) = \frac{1}{2}{(D({P_{up}}) - 1)^2},
\end{equation}
\begin{equation}
{L_{dis}}({P_{up}},{P_T}) = \frac{1}{2}(D{({P_{up}})^2} + {(D({P_T}) - 1)^2}),
\end{equation} where ${P_{up}}$ is the upsampled output of generator $G$, ${P_T}$ is the ground truth point cloud which has the same number of points as ${P_{up}}$. $D({P_{up}})$ and $D({P_T})$ are the predictive confidence value of the discriminator $D$. Eq.(9) can be interpreted as $G$ produces ${P_{up}}$ to confuse $D$ as much as possible by minimizing ${L_{gen}}({P_{up}})$, while Eq.(10) indicates that $D$ needs to learn to distinguish ${P_{up}}$ from ${P_{T}}$ accurately by minimizing ${L_{dis}}({P_{up}},{P_T})$.

\textbf{Reconstruction loss}. To make ${P_{up}}$ fit the underlying surface of ${P_{T}}$ better, we take the Earth Mover's distance (EMD) \cite{b10} as reconstruction loss ${L_{rec}}$ :
\begin{equation}
{L_{rec}} = \mathop {\min }\limits_{\phi :{P_{up}} \to {P_T}} {\sum\nolimits_{{p_i} \in {P_{up}}} {\left\| {{p_i} - \phi ({p_i})} \right\|} _2},
\end{equation}
where ${\phi}$ is a bijection, which encourages point cloud ${P_{up}}$ located close to the underlying surface of ${P_{T}}$. The optimal bijection $\phi :{P_{up}} \to {P_T}$ is unique and invariant under infinitesimal movement of the points \cite{b40}, i.e., for each ${p_i}$ in ${P_{up}}$, the optimal bijection finds the unique point $\phi ({p_i})$ in ${P_{T}}$,  which minimizes the sum of the distances from all ${p_i}$ to $\phi ({p_i})$. As the computation of the optimal bijection is computationally expensive, we use the approximation strategy in \cite{b40} to find a sub-optimal solution in a fixed given time.
%For each point ${p_i}$ in ${P_{up}}$, as there are several corresponding points in the infinitesimal neighbors of ${P_{T}}$, the accurate computation of bijection is too expensive. To reduce %complexity, we adopt approximation strategy in [38] to find an appropriate bijection by allocating fix amount of time for each point and gradually adjusting allowable error for termination.

\textbf{Uniform loss} \cite{b12}. $L_{uni}$ aims to make the points uniformly distributed on the surface of ${P_{up}}$. It contains two parts: global uniformity ${U_{imblance}}({S_i})$ and local uniformity ${U_{cluster}}({S_i})$.
\begin{equation}
{L_{uni}} = \sum\nolimits_{i = 1}^T {{U_{imblance}}({S_i}) \cdot {U_{cluster}}({S_i})},
\end{equation}
where $S_i$, $i = 1,2,..,T$, is a ball queried point subset with radius $r_q$, and $T$ is the number of seed points collected from $P_{up}$ by FPS. The global uniformity ${U_{imblance}}({S_i})$ can be represented as
\begin{equation}
{U_{imblance}}({S_i}) = \frac{{{{(\left| {{S_i}} \right| - \hat n)}^2}}}{{\hat n}},
\end{equation}
where ${\hat n} = rN \times r_q^2$ \cite{b12} ($r$ is the upsampling ratio) indicates the expected number of points in $S_i$. The global uniform regulation ${U_{imblance}}({S_i})$ constrains the number of points in each $S_i$ to be consistent. The local uniformity ${U_{cluster}}({S_i})$ can be written as
\begin{equation}
{U_{cluster}}({S_i}) = \sum\nolimits_j^{\left| {{S_i}} \right|} {\frac{{{{({d_{i,j}} - \hat d)}^2}}}{{\hat d}}},
\end{equation}
where $\hat d = \sqrt {\frac{{2\pi {r_q}^2}}{{\sqrt 3 \left| {{S_i}} \right|}}}$ \cite{b12} refers to the expected uniform distribution distance of points, and ${d_{i,j}}$ is the point-to-nearest-neighbor distance of the $j$-th point in $S_i$. ${U_{cluster}}({S_i})$ forces each point in $S_i$ to have a similar distance from its nearest neighbor, thus ensuring local uniformity.

\textbf{Identity distribution loss}. Point cloud upsampling is essentially an ill-posed problem; therefore, the huge solution space makes it difficult for the network to converge to a global optimum.

Inspired by \cite{b41}, we propose an identity distribution loss $L_{id}$, which reduces the potential solution space by mapping the HR output ${P_{up}} \in$ ${\mathbb{R}^{{rN} \times 3}}$ back to the corresponding LR input ${P_{ori}} \in$ ${\mathbb{R}^{{N} \times 3}}$,

\begin{equation}
{L_{id}} = \mathop {\min }\limits_{\phi :{{\hat P}_{up}} \to {P_{ori}}} {\sum\nolimits_{{{\hat p}_k} \in {{\hat P}_{up}}} {\left\| {{{\hat p}_k} - \phi ({{\hat p}_k})} \right\|}_2},
\end{equation}
where ${{\hat P}_{up}} \in$ ${\mathbb{R}^{{N} \times 3}}$ is an LR point cloud obtained from ${P_{up}}$ by FPS, and ${\phi}$ denotes a bijection between the subsets of equal size ${{\hat P}_{up}}$ and ${P_{ori}}$. $L_{id}$ not only makes $P_{up}$ closer to ${P_{ori}}$ but also effectively reduces the possible solution space of the network to find an optimum quickly.

\textbf{Joint loss}. In summary, we use the generator loss $L_G$ and discriminator loss $L_D$ to train PUFA-GAN, where
\begin{equation}
{L_G} = {w_{gen}}{L_{gen}}({P_{up}}) + {w_{rec}}{L_{rec}} + {w_{uni}}{L_{uni}} + {w_{id}}{L_{id}},
\end{equation}
\begin{equation}
{L_D} = {w_{dis}}{L_{dis}}({P_{up}},{P_T}),
\end{equation}
and ${w_{gen}}$, ${w_{rec}}$, ${w_{uni}}$, ${w_{id}}$ and ${w_{dis}}$ are weights.

\section{EXPERIMENTS}

\subsection{Experimental setup}
\textbf{Datasets}. We used the PU147 dataset in PU-GAN, which includes 147 3D models from the released datasets of PU-Net, MPU, and the Visionair repository \cite{b42}, containing objects of various shapes. Following the protocol in \cite{b12}, 120 point clouds were randomly selected for training, while the rest were used for testing. During the training, 200 patches were randomly cropped from each training point cloud. In total, we collected 24000 patches for training.

\textbf{Network details}. For densely connected feature extraction, the network included four FEBs, where each FEB consists of one DGHRA and two HRA units; the number k of nearest neighbors was 16, and the dimension $C$ was 128. We then used three HRA units connected in cascade in the feature expansion module, and the dimension $C^{\prime}$ of upsampled feature of each HRA was 256. Moreover, we set $N$=256, and $\varepsilon$=0.5. Finally, we empirically set the weight values ${w_{gen}}$, ${w_{rec}}$, ${w_{uni}}$, ${w_{id}}$ and ${w_{dis}}$ in the proposed joint loss to 1, 100, 10, 1, 1, respectively.

%% \multicolumn{2}{c|}
%% \textit{\textbf{$\hat n$ = 1.2 \%}}
%% \bm{\downarrow}
% Table 1
\begin{table*}[htbp]
  \centering
  \begin{threeparttable}
  \caption{Average upsampling accuracy for the test dataset (27 point clouds from \cite{b12}).}
  \renewcommand\arraystretch{1.5}
  \renewcommand\tabcolsep{12pt} % 调整表格列间的宽度
  %\resizebox{\textwidth}{11mm}{
    \begin{tabular}{c|c|c|c|c|c|c}
    \toprule
    \hline
    {\textbf{Method}} & {\textbf{CD}} & {\textbf{HD}} & {\textbf{P2F}} & {\textbf{Uniformity}} & {\textbf{HF\_CD}} & {\textbf{HF\_HD}} \\
    & $\bm{({10^{ - 3}})}$ & $\bm{({10^{ - 3}})}$  & $\bm{({10^{ - 3}})}$ &  $\bm{({10^{ - 3}})}$ & $\bm{({10^{ - 3}})}$ & $\bm{({10^{ - 3}})}$  \\
    \hline
    \textbf{EAR \cite{b6}} & 0.868  & 10.331  & 7.785  &  28.621  &  6.621  & 44.647  \\
    \textbf{PU-Net \cite{b10}} & 0.525  & 5.954  & 13.227  & 974.190  & 3.775  & 25.796  \\
    \textbf{MPU \cite{b11}} & 0.438  & 5.322  & 3.066  & 13.155   & 2.998  & 24.308  \\
    \textbf{PU-Geo \cite{b13}} & 0.405 & 5.338 & 3.432 & 9.259 & 3.111 & 24.778 \\
    \textbf{PU-GAN \cite{b12}} & 0.280  & 4.493  & 2.514  & \textcolor[rgb]{ 1,  0,  0}{\textbf{4.552}} & 2.342 & 22.337 \\
    \textbf{PU-GCN \cite{b15}} & 0.323  & 4.172  & 2.926  & 9.560 & 2.978 & 25.152 \\
    \textbf{Dis-PU \cite{b16}} & 0.289  & 3.734 & \textcolor[rgb]{ 1,  0,  0}{\textbf{2.289}}  & 5.017 & 2.634 & 22.859  \\
    \hline
    \textbf{PUFA-GAN} & \textcolor[rgb]{ 1,  0,  0}{\textbf{0.258}} & \textcolor[rgb]{ 1,  0,  0}{\textbf{3.571}} & \textcolor[rgb]{ .337,  .29,  .973}{\textbf{2.392 }} & \textcolor[rgb]{ .337,  .29,  .973}{\textbf{4.889 }} &  \textcolor[rgb]{ 1,  0,  0}{\textbf{2.081}} & \textcolor[rgb]{ 1,  0,  0}{\textbf{19.744}} \\
    \hline
    \bottomrule
    \end{tabular}%}
    \begin{tablenotes}
    % \footnotesize
    \item *Note that the lower the evaluation value, the better the performance. The bold red and blue values represent best and second best results, respectively.
    \end{tablenotes}
    \end{threeparttable}
  \label{tab:addlabel}%
  \vspace{-1.5em}
\end{table*}%

\begin{figure*}[!htp]
\centering
\includegraphics[width=18.8cm,height=9.3cm]{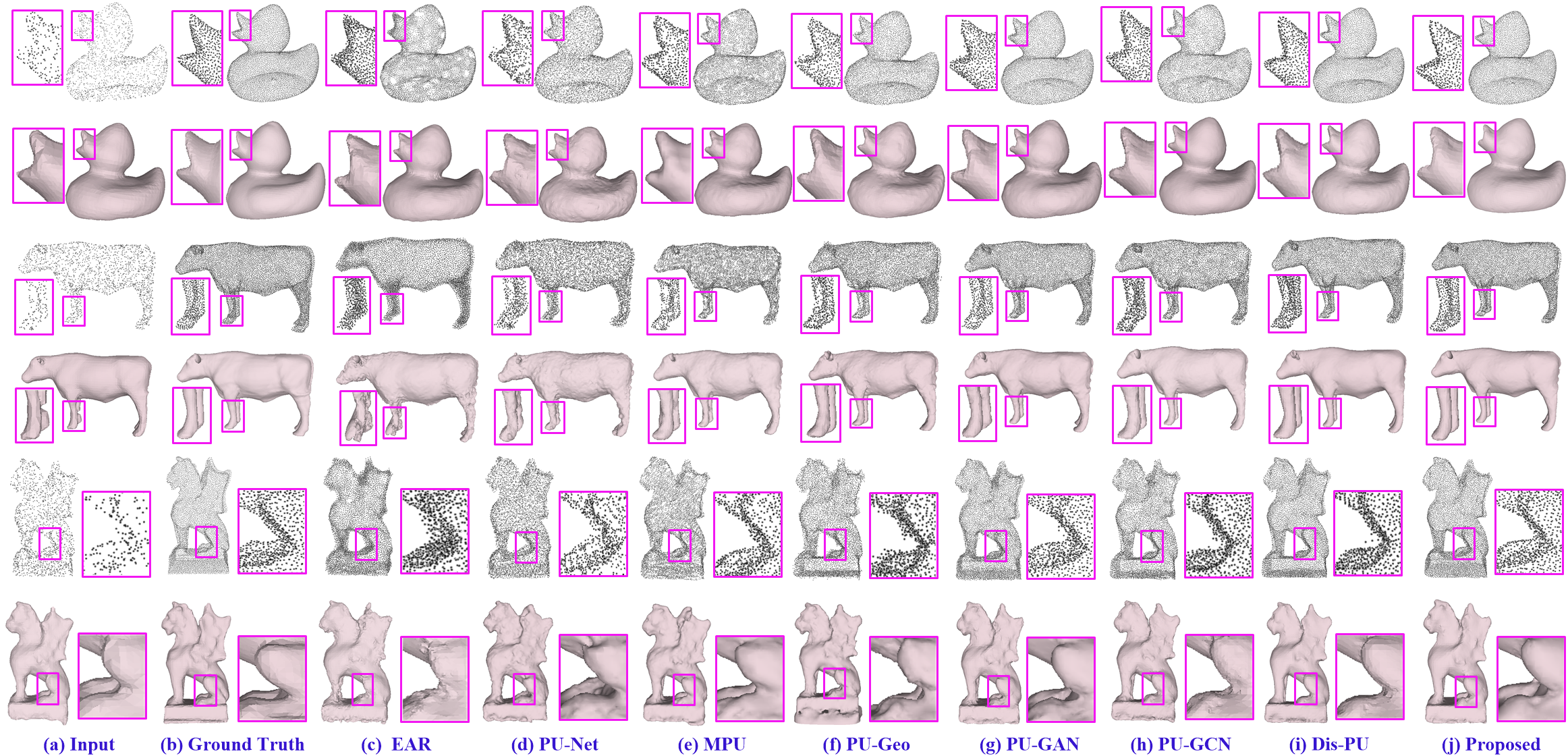}
\caption{Upsampling ($\bm{\times}$4) and surface reconstruction results produced with all methods (c)-(j) from 2048 input points (a).}
\vspace{-1.5em}
\end{figure*}

\textbf{Training}. To enhance the robustness of the network, we used the following data enhancement strategies for input patches: point perturbation with Gaussian noise and random rotation and scaling. We trained the proposed PUFA-GAN for 130 epochs with a batch size of 28. The Adam algorithm \cite{b43} with a two time-scale update rule (TTUR) \cite{b44} was used, and the learning rate of the generator and discriminator were 0.0009 and 0.0006, respectively. We implemented the proposed method in the Tensorflow platform. A computer with an Intel Core i7 7820X processor, an NVIDIA Tesla V100 GPU, and 64GB memory was used to conduct the experiments.

\textbf{Testing}. The patch-fusion strategy in \cite{b11} is used to generate the ultimate upsampled point cloud. Specifically, a series of seed points are first picked by FPS for each test sample. Then, for each seed point, a local patch with 256 points is generated. Next, all patches are fed into the generator and all the upsampled patches are fused together. Finally, the ultimate upsampled point cloud with the specified upsampling ratio is obtained by FPS from the fused patches. However, subjective experiments showed (see Fig.1) that there may be noise in HF regions due to the overlap between upsampled patches before patch fusion. To alleviate this problem, we propose to use another efficient graph filter to explicitly remove any potential high-frequency noise from each upsampled patch before patch fusion, which is unique to our method. In this way, we can effectively eliminate noise and retain more detailed textures in HF regions.

\subsection{Evaluation metrics}

To assess the upsampling performance quantitatively, we used four common evaluation metrics: \textbf{Chamfer distance (CD)} \cite{b40}, \textbf{Hausdorff distance (HD)} \cite{b45}, \textbf{Point-to-surface (P2F) distance} \cite{b46} and \textbf{Uniformity} \cite{b12}. \textbf{CD} and \textbf{HD} measure the shape difference between $P_{up}$ and $P_{T}$ with the Euclidean distance; \textbf{P2F} assesses the deviation of $P_{up}$ from the underlying surface of $P_{T}$; the \textbf{Uniformity} metric evaluates the uniformity of $P_{up}$, and we set ${r_q}^2 = 0.012$ (see Uniform loss in Section 3.4). Because existing metrics cannot effectively appraise the difference in HF regions (which are usually more prone to noise) between $P_{up}$ and $P_{T}$, we also propose two metrics to evaluate the regularity in HF regions. Specifically, the HF points of the upsampled point cloud $P_{up}$ and ground truth $P_{T}$ are first extracted by GF. Then, the first metric, which we call \textbf{HF\_CD}, is obtained by calculating the \textbf{CD} of the corresponding HF points, i.e.,
\begin{equation}
\begin{aligned}
HF\_CD = \frac{1}{M}(\sum\nolimits_{p \in {P_{up\_c}}} {\mathop {\min }\limits_{q \in {P_{T\_c}}} \left\| {p - q} \right\|_2^2 + } \\
\sum\nolimits_{q \in {P_{T\_c}}} {\mathop {\min }\limits_{p \in {P_{up\_c}}} \left\| {q - p} \right\|_2^2)},
\end{aligned}
\end{equation}
and the second metric, which we call \textbf{HF\_HD}, is obtained by calculating the \textbf{HD} of the corresponding HF points, i.e.,
\begin{equation}
\begin{aligned}
HF\_HD = \max (\mathop {\max }\limits_{p \in {P_{up\_c}}} \mathop {\min }\limits_{q \in {P_{T\_c}}} {\left\| {p - q} \right\|_2}, \\
\mathop {\max }\limits_{q \in {P_{T\_c}}} \mathop {\min }\limits_{p \in {P_{up\_c}}} {\left\| {q - p} \right\|_2}),
\end{aligned}
\end{equation}
Here, ${P_{up\_c}}$ and ${P_{T\_c}}$ are the HF points of $P_{up}$ and $P_{T}$, respectively, $p$ and $q$ represent two points in ${P_{up\_c}}$ and ${P_{T\_c}}$, respectively, and $M$ is the number of HF points.

% Table 2
\begin{table*}[htbp]
  \setlength{\abovecaptionskip}{0cm}  %段前
  \setlength{\belowcaptionskip}{-0.2cm} %段后
  \centering
  \caption{Train and inference time.}
  \renewcommand\arraystretch{1.5}
  \renewcommand\tabcolsep{11pt} % 调整表格列间的宽度
    \begin{tabular}{c|c|c|c|c|c|c|c}
    \toprule
    \hline
    {\textbf{Time}} & {\textbf{PU-Net}} & {\textbf{MPU}} & {\textbf{PU-Geo}} & {\textbf{PU-GAN}} & {\textbf{PU-GCN}} & {\textbf{Dis-PU}} & {\textbf{PUFA-GAN}} \\
    \hline
    \textbf{Train (h)} &  5  &  6  &  5  & 18  &  2  &  23  &  28  \\
    \hline
    \textbf{Inference (s)} & 0.13  & 1.06  & 0.7  & 0.46  &  0.26  & 0.58 & 0.94\\
    \hline
    \bottomrule
    \end{tabular}%
  \label{tab:addlabel}%
  \vspace{-1.1 em}
\end{table*}%

% Table 3
\begin{table*}[htbp]
  \setlength{\abovecaptionskip}{0cm}  %段前
  \setlength{\belowcaptionskip}{-0.2cm} %段后
  \centering
  \caption{Upsampling ($\bm{\times}$4) results on ModelNet40.}
  \renewcommand\arraystretch{1.5}
  \renewcommand\tabcolsep{12pt} % 调整表格列间的宽度
    \begin{tabular}{c|c|c|c|c|c|c}
    \toprule
    \hline
     {\textbf{Method}} & {\textbf{CD}} & {\textbf{HD}} & {\textbf{P2F}} & {\textbf{Uniformity}} & {\textbf{HF\_CD}} & {\textbf{HF\_HD}} \\
    & $\bm{({10^{ - 3}})}$ & $\bm{({10^{ - 3}})}$  & $\bm{({10^{ - 3}})}$ &  $\bm{({10^{ - 2}})}$ & $\bm{({10^{ - 3}})}$ & $\bm{({10^{ - 3}})}$  \\
    \hline
    \textbf{EAR \cite{b6}} & 1.332  & 12.284  & 6.899 & 65.006 & 10.923 & 96.297  \\
    \textbf{PU-Net \cite{b10}} & 1.189  & 11.923  & 7.103 & 33.581  & 5.596 & 82.237  \\
    \textbf{MPU \cite{b11}} & 0.910  & 8.604  & 2.799  & 36.425  & 5.370  & 86.518  \\
    \textbf{PU-Geo \cite{b13}} & 0.576  & 7.088  & 2.828 & 32.931 & 3.246 & 75.551 \\
    \textbf{PU-GAN \cite{b12}} & 0.401  & 6.662  & 2.392 & 22.863  & 2.586  & 66.254  \\
    \textbf{PU-GCN \cite{b15}} & 0.586  & 6.539  & 2.586 & 34.937  & 3.803  & 76.215  \\
    \textbf{Dis-PU \cite{b16}} & 0.537  & 6.196  & 2.315 & 25.468  & 2.722  & 60.590  \\
     \hline
    \textbf{PUFA-GAN} & \textcolor[rgb]{ 1,  0,  0}{\textbf{0.379 }} & \textcolor[rgb]{ 1,  0,  0}{\textbf{5.891 }} & \textcolor[rgb]{ 1,  0,  0}{\textbf{2.170 }} & \textcolor[rgb]{ 1,  0,  0}{\textbf{17.806}}  &\textcolor[rgb]{ 1,  0,  0}{\textbf{2.544 }} & \textcolor[rgb]{ 1,  0,  0}{\textbf{59.823 }} \\
    \hline
    \bottomrule
    \end{tabular}%
  \label{tab:addlabel}%
  \vspace{-0.5em}
\end{table*}%

\begin{figure*}[!htp]
\centering
\includegraphics[width=18.5cm,height=3.1cm]{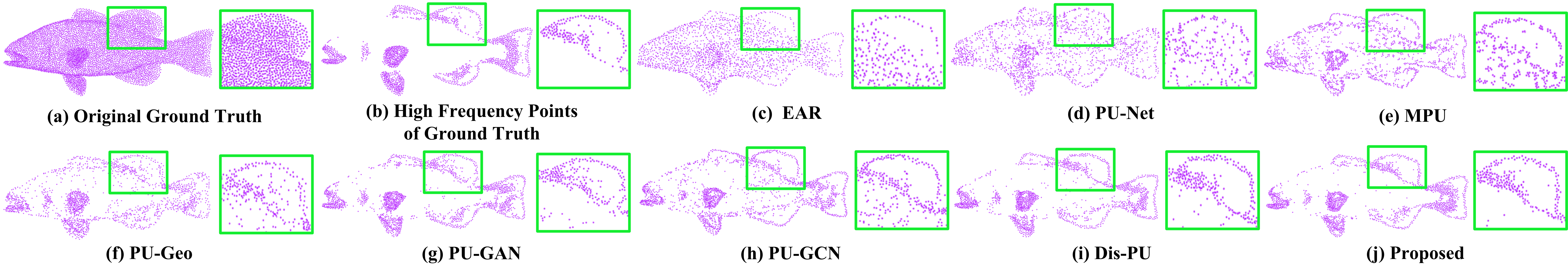}
\caption{Upsampling ($\bm{\times}$4) results of high frequency points of \textit{m60} from the PU147 dataset.}
\vspace{-1.2em}
\end{figure*}

In the computation of \textbf{CD}, \textbf{HD}, \textbf{P2F}, and \textbf{Uniformity}, the ground truth point cloud $P_{T}$ was obtained by sampling 8192 points from the test point cloud using Poisson disk sampling. The LR point cloud was obtained by selecting 2048 points from the test point cloud using Monte Carlo random sampling. In the computation of \textbf{HF\_CD} and \textbf{HF\_HD}, both ${P_{up\_c}}$ and ${P_{T\_c}}$ consisted of 2048 points obtained from $P_{up}$ and $P_{T}$ with GF.

\subsection{Comparison with the state-of-the-art}
We compared PUFA-GAN with seven methods: optimization based EAR \cite{b6}, and six state-of-the-art learning-based methods: PU-Net \cite{b10}, MPU \cite{b11}, PU-GAN \cite{b12}, PU-Geo \cite{b13}, PU-GCN \cite{b15} and Dis-PU \cite{b16}. For a fair comparison, we re-trained PU-Net, MPU, PU-GAN, PU-GCN, and Dis-PU on our training dataset. This was not possible for PU-Geo because its training requires the normal vector information, which is not included in our training dataset. As a compromise, we used the trained model which was provided to us by the corresponding author \cite{b13}. The upsampling ratio $r$ was set to 4.

\textbf{Quantitative results}. Table 1 shows the quantitative upsampling performance of all the methods. We can see that PUFA-GAN performs best in terms of \textbf{CD} (0.258), \textbf{HD} (3.571), \textbf{HF\_CD} (2.081), and \textbf{HF\_HD} (19.744). For the \textbf{P2F} and \textbf{Uniformity} metrics, PUFA-GAN obtained the second best results.

\textbf{Qualitative results}. In addition to the objective comparison, we also show in Fig. 6 the subjective comparisons of upsamping and surface reconstruction \cite{b47}. We can see that the outputs of EAR, PU-Net and MPU are significantly unevenly distributed. Some local holes may also appear. For PU-Geo, PU-GAN, PU-GCN and Dis-PU, although the generated point clouds are uniform, there are noisy points in HF regions. Compared with these methods, PUFA-GAN not only generates smoother point clouds attached to the underlying surface but also forms a clear edge with fine-grained details. Specifically, the mouth of $\bm{duck}$ (top rows) shows that our method generates a uniform and neat edge, while other methods fail. Moreover, the zoomed-in views show that our method preserves more detailed texture, e.g., $\bm{cow}$'s hoof (middle rows) and $\bm{statue}$'s leg (bottom rows). Besides, we also visualize the HF points of an example fish named $\bm{m60}$ in the PU147 dataset for all methods (Fig. 7). The same high-pass graph filter was used to extract the HF points. Compared with the other methods, PUFA-GAN generated a cleaner HF region, where the points are closer to the HF points of the ground truth. More visual comparisons can be found in the \textbf{supplementary materials}.

\textbf{Training and inference time comparison}. The time for train and inference are compared in Table 2. Note that EAR is an optimization-based method, and for each sparse input point cloud, we need to constantly adjust various hyper-parameters to obtain the final dense point cloud. Therefore, we do not report its CPU time.  As shown in Table 2, the training time of PU-GCN was the shortest (2 h), while the inference time of PU-Net was the shortest (0.13 s). With our method, the training time and the inference time were 28 h and 0.94 s, respectively. The higher time complexity compared to the previous methods is mainly due to the use of the GF for extracting the high frequency points.

\begin{figure*}[!htp]
\centering
\includegraphics[width=17cm,height=4cm]{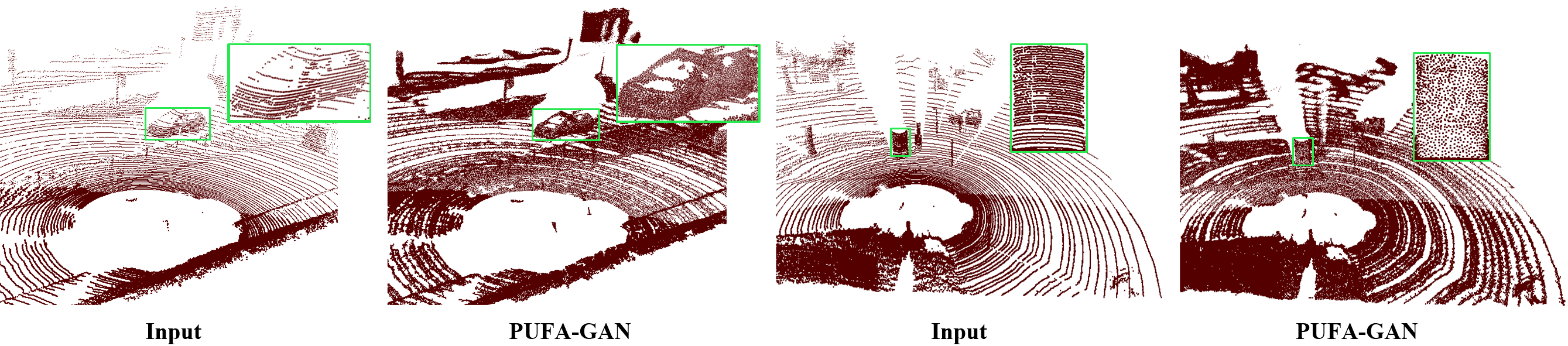}
\caption{Upsampling ($\bm{\times}$4) the real scanned LiDAR point clouds with zoomed-in views (Car and Wall) from the KITTI dataset with PUFA-GAN.}
\vspace{-0.7em}
\end{figure*}

\begin{figure*}[!htp]
\centering
\includegraphics[width=15cm,height=4.4cm]{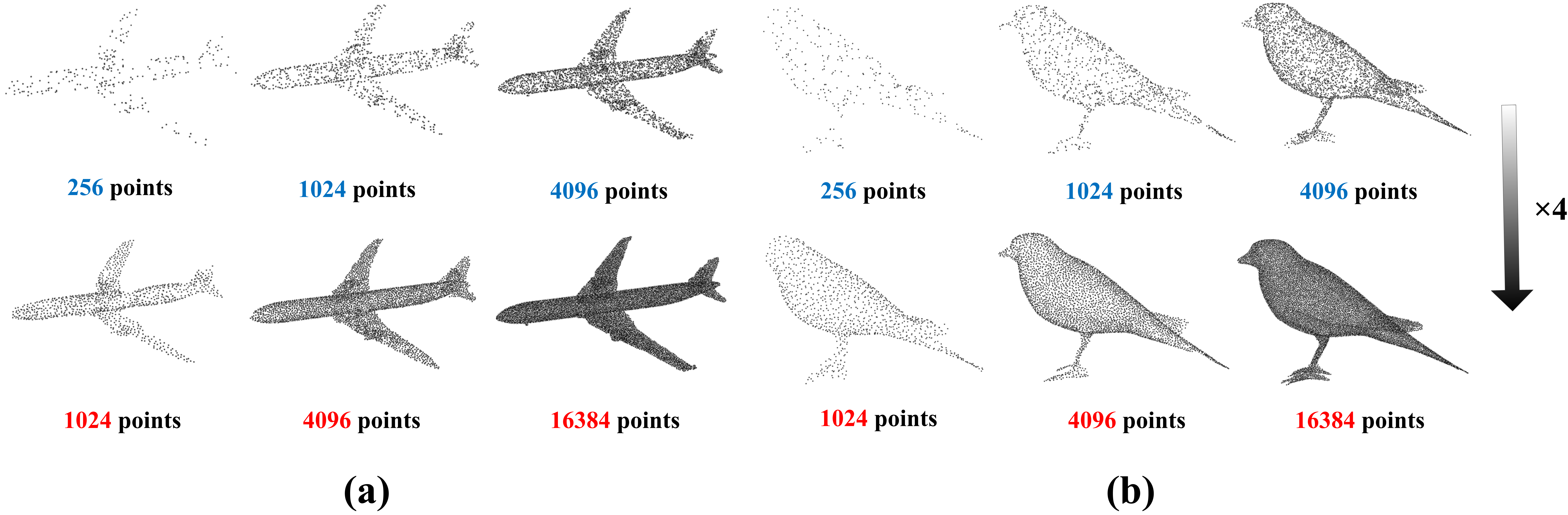}
\caption{Visual upsampling ($\bm{\times}$4) results of the unseen \textit{Airplane} (a) and test point cloud \textit{Bird} (b) with input sizes 256, 1024 and 4096.}
\vspace{-0.7em}
\end{figure*}

\begin{figure*}[!htp]
\centering
\includegraphics[width=14.5cm,height=6cm]{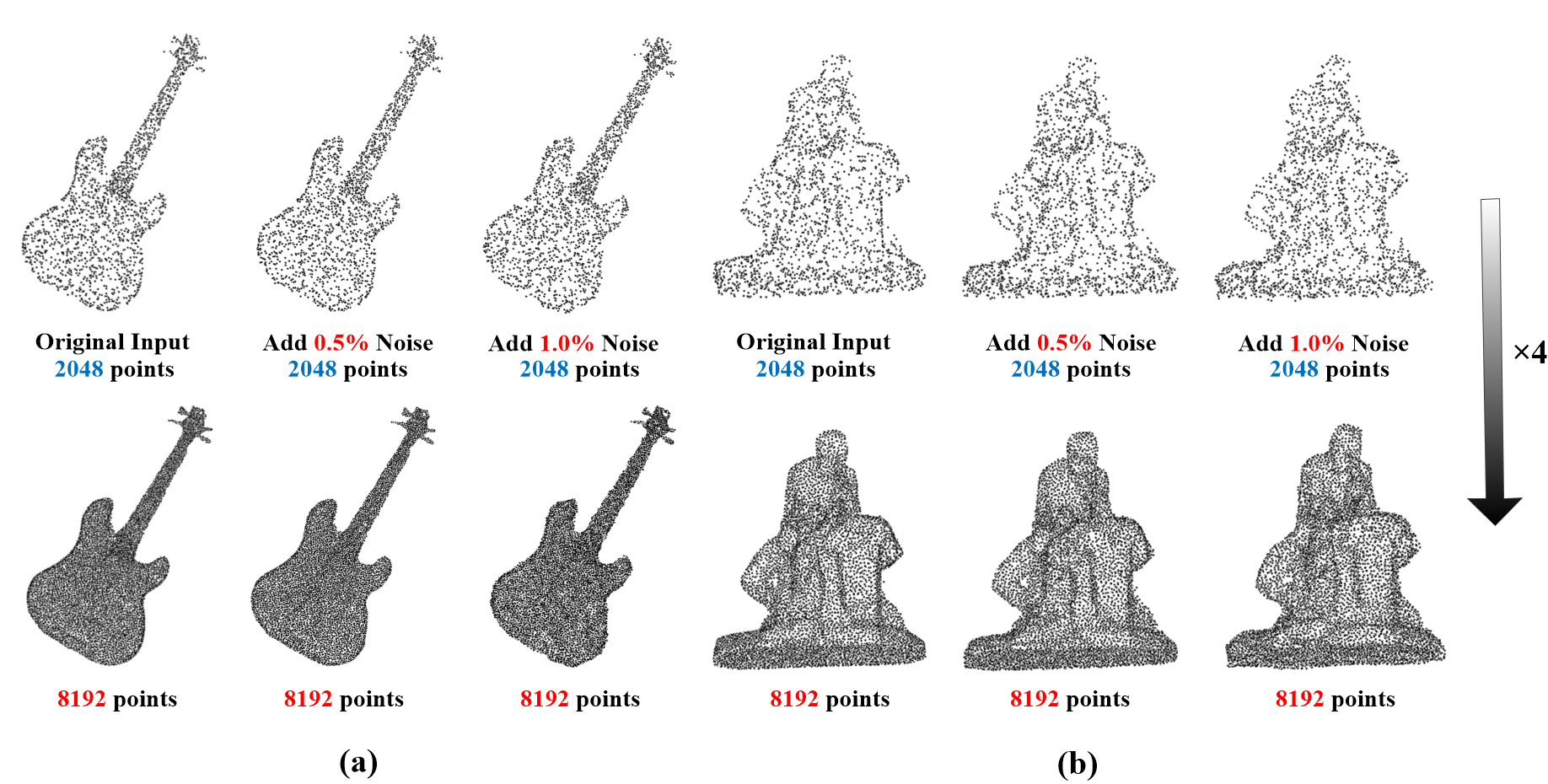}
\caption{Using PUFA-GAN to upsample ($\bm{\times}$4) noisy point clouds (input point clouds with Gaussian noise levels: 0, 0.5\%, and 1\% from left to right). Note that the unseen \textit{Guitar} (a) is selected from ModelNet40, and \textit{Sculpture} belongs to our test dataset.}
\vspace{-1.5em}
\end{figure*}
\vspace{-0.8em}
\subsection{Robustness Analysis}
We further evaluated the robustness of PUFA-GAN from the following aspects: unseen (test directly on a new dataset using a trained model) point clouds, real scanned large-scale LiDAR point clouds, point clouds with various sizes and noise.

\textbf{Upsampling unseen point clouds}. To demonstrate the generalization ability of PUFA-GAN, we tested it on the unseen ModelNet40 \cite{b48}. We randomly selected 20 point clouds from different categories in ModelNet40. The ground truth of each test sample containing 8192 points was obtained by the Poisson disk sampling algorithm. At the same time, Monte-Carlo random sampling was used to acquire the test input with 2048 points. As shown in Table 3, compared with the other methods, PUFA-GAN achieved the best performance for all evaluation metrics.

\textbf{Upsampling real scanned LiDAR point clouds}. In general, real scanned LiDAR point clouds are sparse and include a lot of noise and occlusions; thus it is challenging to upsample them precisely. We verified the effectiveness of PUFA-GAN on the real scanned LiDAR point clouds from the KITTI \cite{b49} dataset (Fig. 8). We can see that PUFA-GAN can generate dense and uniform scenes with rich geometric details.

\textbf{Upsampling point clouds with varying sizes}. We evaluated PUFA-GAN on two different datasets (ModelNet40 and our test datasets) for input point clouds of different scale. The Monte-Carlo random sampling algorithm was adopted to sample the test point clouds to input sizes 256, 1024 and 4096. Fig. 9 shows that even for only 256 points, our method can recover the approximate shapes of the sparse input point clouds.

% Table 4
\begin{table*}[htbp]
  \setlength{\abovecaptionskip}{0cm}  %段前
  \setlength{\belowcaptionskip}{-0.2cm} %段后
  \centering
  \caption{Ablation study on the PU147 dataset.}
  \renewcommand\arraystretch{1.5}
  \renewcommand\tabcolsep{11pt} % 调整表格列间的宽度
    \begin{tabular}{c|c|c|c|c|c|c}
    \toprule
    \hline
   {\textbf{Method}} & {\textbf{CD}} & {\textbf{HD}} & {\textbf{P2F}} & {\textbf{Uniformity}} & {\textbf{HF\_CD}} & {\textbf{HF\_HD}}  \\
     & $\bm{({10^{ - 3}})} $ & $\bm{({10^{ - 3}})} $  & $\bm{({10^{ - 3}})} $ &  $\bm{({10^{ - 3}})} $ & $\bm{({10^{ - 3}})} $ & $\bm{({10^{ - 3}})} $  \\
   \hline
    \textbf{PUFA-GAN w/o DGHRA} & 0.279  & 4.211  & 2.715  & 6.705  & 2.579  & 21.178  \\
    \textbf{PUFA-GAN w/o HRA} & 0.482  & 14.601  & 4.789  & 15.220  & 4.582  & 33.253  \\
    \textbf{PUFA-GAN w/o GF} & 0.258  & 3.862  & 2.408  & 4.898  & \textcolor[rgb]{ 1,  0,  0}{\textbf{2.044 }} & 21.024  \\
    \textbf{PUFA-GAN w/o HFD} & 0.269  & 4.059  & 2.428  & 5.374  & 2.207  & 20.301  \\
    \textbf{PUFA-GAN w/o IDL} & 0.266  & 4.029  & 2.403  & 5.781  & 2.187  & 20.788  \\
    \hline
    \textbf{PUFA-GAN} & \textcolor[rgb]{ 1,  0,  0}{\textbf{0.258 }} & \textcolor[rgb]{ 1,  0,  0}{\textbf{3.571 }} & \textcolor[rgb]{ 1,  0,  0}{\textbf{2.392 }} &  \textcolor[rgb]{ 1,  0,  0}{\textbf{4.889 }} & \textcolor[rgb]{ .337,  .29,  .973}{\textbf{2.081 }} & \textcolor[rgb]{ 1,  0,  0}{\textbf{19.744 }} \\
    \hline
    \bottomrule
    \end{tabular}%
  \label{tab:addlabel}%
  \vspace{-1.5em}
\end{table*}%

\begin{figure}[!htp]
\centering
\includegraphics[width=8.7cm,height=5.2cm]{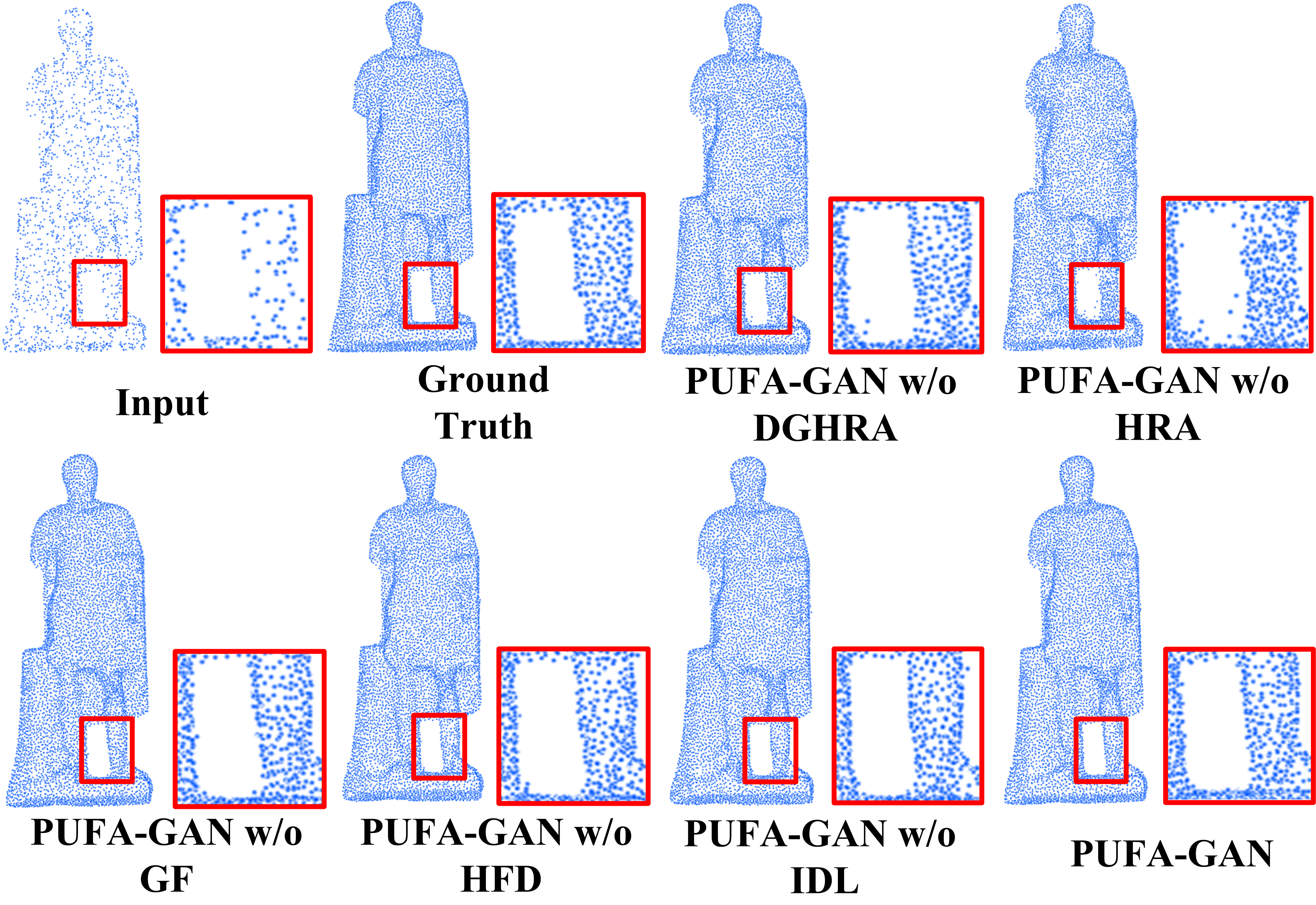}
\caption{Ablation study visual results ($\bm{\times}$4).}
\end{figure}

\textbf{Upsampling noisy point clouds}. To demonstrate the anti-noise ability of PUFA-GAN, we upsampled the unseen (from ModelNet40) and test point clouds with Gaussian noise at different noise levels. Fig. 10 shows that for the gradually enhanced noise, our method can also preserve the natural output shape and avoids noise interference.

\subsection{Ablation Study}
We conducted an ablation study to quantitatively and qualitatively evaluate the performance of each proposed component, i.e., 1) the densely connected feature extraction module, 2) the hierarchical residual-based feature expasion module, 3) the GF unit in the patch-fusion strategy, 4) the frequency-aware discriminator (FAD), and 5) the identity distribution loss (IDL). In the study, the following five networks were considered.

1) \textbf{PUFA-GAN w/o DGHRA}: the proposed feature extraction module is replaced with DGCNN \cite{b29}.

2) \textbf{PUFA-GAN w/o HRA}: HRA is replaced with the same number of MLPs in the feature expansion module.

3) \textbf{PUFA-GAN w/o GF}: the GF is not used in the patch-fusion strategy.

4) \textbf{PUFA-GAN w/o HFD}: the HF discriminator (HFD) is removed from FAD (i.e., only a global discriminator is kept)

5) \textbf{PUFA-GAN w/o IDL}: IDL is removed from the joint loss.

The results in Table 4 and Fig. 11 show that PUFA-GAN with all the proposed components performs best. When any component was removed, the performance of the network decreased. Particularly, we can see that the most dramatic performance degradation occured when the HRA unit in the feature expansion module was removed, which is in line with our expectations. This is because the feature expansion module directly determines the quality of the upsampling features with the help of the hierarchical residual features, while the naive MLPs do not have the ability to explore the feature details. On the other hand, the GF in the patch-fusion strategy had the weakest influence, since it can be regarded as a simple post-processing step and does not affect the entire performance of PUFA-GAN.

\section{Conclusion}
We presented a novel generative adversarial network for point cloud upsampling, namely PUFA-GAN. The generator includes three modules: feature extraction, feature expansion, and geometry generation. Specifically, a densely connected DGHRA-based feature extraction module is proposed to obtain point-wise descriptive features. Then an efficient cascaded HRA-based feature expansion module generates and enriches the feature details. Next, the upsampled features are mapped back to the geometry domain in the geometry generation module. To improve the quality of the upsampled point clouds, we proposed a frequency-aware discriminator which consists of a global discriminator and an HF discriminator. The former considers the global context for better skeleton recognition, while the latter suppresses noise in HF regions. To further produce clean local details, we removed the potential noise explicitly by GF in the testing phase. In addition, we proposed a novel identity distribution loss function to force LR input point clouds and HR output point clouds to have identical geometry distributions. Finally, we introduced two evaluation metrics (\textbf{HF\_CD} and \textbf{HF\_HD}) to evaluate the regularity of upsampled point clouds in HF regions quantitatively. Extensive experiments demonstrated that PUFA-GAN outperforms state-of-the-art methods.

In the future, we will design a network with an arbitrary upsampling ratio for flexible scene switching in an end-to-end fashion. We will also explore other research directions, such as i) upsampling point clouds with attribute information (e.g., color, reflectance, normal vector) instead of only geometry information; ii) taking the point cloud upsampling as a point cloud completion problem; iii) adopting unsupervised or semi-supervised learning to alleviate the dependence on paired and scarce training samples.

%\appendices
%\section{Graph Filter Theory}

\end{document}